\RequirePackage{fix-cm}
\documentclass[twocolumn]{svjour3}          
\smartqed  
\usepackage{graphicx}

\usepackage{times}
\usepackage{epsfig}
\usepackage{amsmath}
\usepackage{amssymb}

\usepackage{booktabs}
\usepackage{multirow}
\usepackage{tabularx}

\journalname{VRIH} 

\newcommand{\YL}{}

\begin{document}

\title{Generating Animatable 3D Cartoon Faces from Single Portraits}
\subtitle{}
\author{Chuanyu Pan \and Guowei Yang \and Taijiang Mu \and Yu-Kun Lai}
\institute{Chuanyu Pan \at University of California Berkeley \and Guowei Yang, Taijiang Mu \at Tsinghua University \and Yu-Kun Lai \at Cardiff University}
\date{ }

\maketitle

\begin{abstract}
With the booming of virtual reality (VR) technology, there is a growing need for customized 3D avatars. However, traditional methods for 3D avatar modeling are either time-consuming or fail to retain similarity to the person being modeled. We present a novel framework to generate animatable 3D cartoon faces from a single portrait image. We first transfer an input real-world portrait to a stylized cartoon image with a  StyleGAN. Then we propose a two-stage reconstruction method to recover the 3D cartoon face with detailed texture, which first makes a coarse estimation based on template models, and then refines the model by non-rigid deformation under landmark supervision. Finally, we propose a semantic preserving face rigging method based on manually created templates and deformation transfer. Compared with prior arts, qualitative and quantitative results show that our method achieves better accuracy, aesthetics, and similarity criteria. Furthermore, we demonstrate the capability of real-time facial animation of our 3D model.
\end{abstract}


\section{Introduction}

\begin{figure*}[t]
\begin{center}
\includegraphics[width=1.0\linewidth]{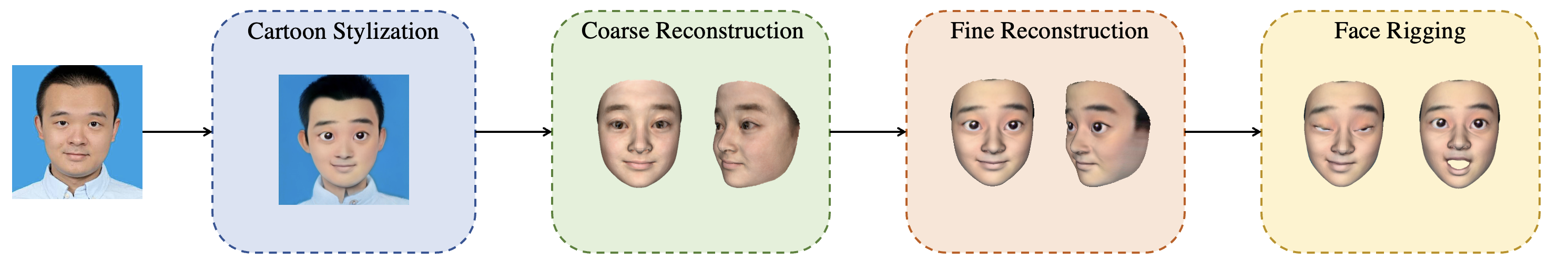}
\end{center}
   \caption{The pipeline of our animatable 3D cartoon face \YL{generation} method. We first transform an input portrait to a 2D cartoon image. Then we conduct template-based coarse reconstruction and deformation-based fine reconstruction to build an elaborate 3D cartoon face. Finally, we generate semantic face rigs for facial animation, making the static 3D model animatable.}
\label{fig:pipeline}
\end{figure*}

\YL{Virtual and augmented reality (VR/AR)} has developed rapidly in recent years. An essential and challenging task in this field is to create virtual 3D faces for users and avatars. These faces should \YL{achieve} high performance on aesthetics and \YL{recognizability, resembling the person being modeled.} 
\YL{They} should also be animatable \YL{for many downstream applications}. However, traditional methods either require heavy manual modeling, \YL{which is time consuming}, or rely on existing general templates, \YL{and thus tend to lose recognizability}. With the development of deep learning techniques, a few methods~\cite{feng2018joint,deng2019accurate,guo2018cnn} have been proposed that automatically reconstruct \YL{realistic 3D} faces from images. However, due to numerous facial details, realistic 3D reconstruction becomes extremely hard to reach high similarity with the reference face. In comparison, cartoon faces are easier to reach high visual performance and can be represented with less memory. Therefore, many VR/AR applications choose 3D cartoon faces \YL{as avatars} for user images.  

Our work \YL{focuses} on automatically creating 3D animatable cartoon \YL{faces} based on a single real-world portrait. As Fig.~\ref{fig:pipeline} shows, we split our pipeline into the following \YL{steps}: we first generate a stylized cartoon image from the input image with a StyleGAN~\cite{karras2019style}; then we reconstruct a static 3D cartoon face from the stylized image; finally, we generate semantic-preserving facial rigs to make the 3D face animatable. 

Existing face reconstruction methods~\cite{feng2018joint,deng2019accurate} perform poorly in reconstructing cartoon faces because they introduce strong real-world \YL{priors} that \YL{are} hard to generalize to the cartoon domain. Some works~\cite{qiu20213dcaricshop} that reconstruct 3D caricatures fail to perform \YL{well} on real-world 
\YL{portrait images}
due to domain \YL{gaps} as well. However, to \YL{obtain} accurate texture mapping and natural facial animation, precise correspondences between the reconstructed 3D face and semantic labels on the 2D image are required. These correspondences are usually acquired by projecting the model back to the image. Therefore, wrong shapes would cause wrong correspondences, \YL{highlighting the necessity for accurate reconstruction} in this task.

To solve this problem, we propose a two-stage reconstruction method. In the first stage, we utilize face templates and a reconstruction network to make a coarse estimation. In the second stage, our non-rigid deformation refinement adjusts the 3D model under the supervision of accurate 2D annotations. This refinement is not restricted to 
\YL{a specific domain}.
Some works~\cite{guo2018cnn,cao2015real} introduced a similar idea of adding a \YL{refinement} network to adjust the 3D model. However, these works constrain the refinement on depth or normal directions. \YL{As a result, they are} effective in reconstructing face details like wrinkles and moles. But for cartoon faces, which usually contain larger eyes and exaggerated expressions, these refinements \YL{are insufficient to handle}. Our method conducts 
\YL{more general}
refinement, making accurate alignment without unnatural distortions. We show that our method performs well on both cartoon and real-world data.

Face rigging is the last part of our pipeline, which is the \YL{basis for} facial animation. Facial animation methods~\cite{blanz2003reanimating} that use 3D morphable models (3DMM)~\cite{blanz1999morphable} usually \YL{lack} semantics, making  \YL{it} hard to apply \YL{them} to industrial applications. Some face rigging methods~\cite{li2010example,zhou20183d} can generate semantic rigs but require user-specific training samples. Our semantic-preserving rigging method \YL{conducts} deformation transfer from a set of hand-made expression models to the target. The expression models are predefined and built by professional modelers, and the rigging process is free from any reference samples. 

Our work is industry-oriented, aiming to realize high-quality customized cartoon face reconstruction with real-time animation capability. Experiments show that our method outperforms prior arts on both reconstruction accuracy and user subjective evaluation. We show visualization results and an application of real-time 
\YL{video driven}
animation. In summary, our main contributions are:

\begin{enumerate}
    \item \YL{We develop} a complete system that generates a user-specific 3D cartoon face from a single portrait, which is real-time animatable. It can be directly applied to VR/AR applications such as virtual meetings and social networking for 
    \YL{avatar} customization.
    \item \YL{To achieve this, we propose} a two-stage 3D face reconstruction scheme that \YL{produces} high-quality results on both real-world \YL{portrait images} and cartoon \YL{images}. Our deformation-based refinement in the second stage evidently improves the performance of texture mapping and facial animation. 
    \item \YL{We further provide} a solution for semantic-preserving face rigging without reference samples.
\end{enumerate}
\begin{figure*}[t]
\begin{center}
\includegraphics[width=1.0\linewidth]{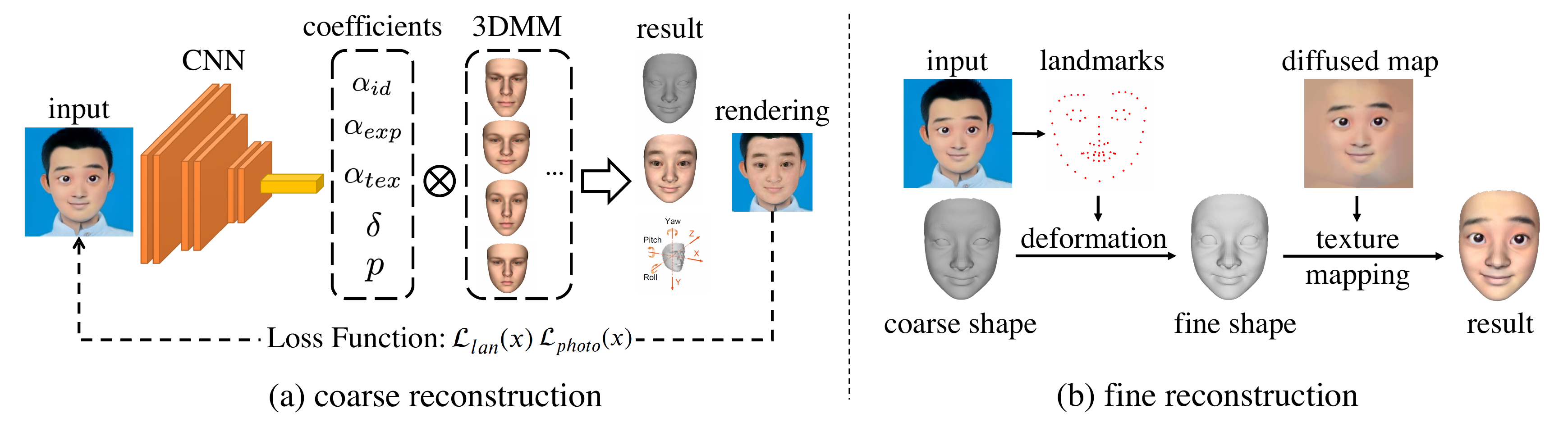}
\end{center}
   \caption{Overview of our two-stage 3D cartoon face reconstruction. (a) Our coarse reconstruction method utilizes a CNN to predict 3DMM coefficients from an input image. The output coefficients contain \YL{a} combination \YL{of} parameters \YL{for} identity $\alpha_{id}$, expression $\alpha_{id}$, texture $\alpha_{tex}$, lighting $\delta$, and pose $p$. (b) Our fine reconstruction method \YL{refines} the coarse shape using landmark supervision with \YL{Laplacian} deformation. The refined model is then colored by diffused texture.}
\label{fig:recon_pipeline}
\end{figure*}

\section{Related Work}

\paragraph{Model-based Single Image 3D Face Reconstruction.} 3D \YL{face reconstruction} has been studied extensively in 3D computer vision, which is \YL{widely} applied in face recognition, character generation, facial data collection, etc. The reconstructed 3D faces are usually represented \YL{as} 3D meshes with a large number of vertices. To reduce the complexity of face representation, 3D Morphable Models (3DMM)~\cite{blanz1999morphable} have been proposed for face modeling. 3DMM is a set of basis that constructs a low-dimensional subspace of 3D faces. The geometry and texture of the faces that \YL{reside in} the manifold can be expressed by linear \YL{combinations} of the basis. \YL{Some works}~\cite{blanz2004statistical,zhu2015high,hassner2015effective,bas2016fitting} align the reconstructed face model with facial landmarks on the input image to regress 3DMM coefficients. However, these methods have difficulties capturing detailed geometry of the faces due to the landmark sparsity. Other works use features like image \YL{intensities} and edges~\cite{romdhani2005estimating} to preserve facial fidelity. With the development of deep learning and differentiable rendering, some recent works~\cite{kim2017inversefacenet,jourabloo2016large,zhu2016face} use \YL{Convolutional Neural Networks (CNNs)} to learn the 3DMM coefficients and pose parameters. To alleviate the lack of training data, \YL{Deng et al.}~\cite{deng2019accurate} utilize photometric information to train CNNs in a weakly-supervised manner. All these 3DMM-based methods are facing the same problem: exaggerated shapes and geometry details can hardly be preserved due to the lack of expressivity of the low-dimensional \YL{linear} models. To tackle this problem, \YL{Guo et al.}~\cite{guo2018cnn} propose a \YL{finetuning} network to recover geometry details, such as wrinkles and moles, after 3DMM coarse reconstruction. However, this method restricts the \YL{finetuning} displacement to the depth direction, \YL{so} still not capable of reconstructing exaggerated expressions and shapes, like large eyes and big mouth, which are quite common in cartoon images. There are also some model-free single-image reconstruction methods~\cite{hassner2006example,kemelmacher2011face,hassner2013viewing,feng2018joint}. However, the results of these methods are hard to be aligned or animated due to the topological inconsistency of the output meshes. 

\paragraph{Stylized Face Reconstruction.} Stylized faces usually come with larger variation of shapes and expressions, making it difficult to transfer realistic reconstruction methods to the cartoon domain directly. \YL{Liu et al.}~\cite{liu2009semi} represent \YL{3D} caricatures with 3DMM. Since 3DMM is low-dimensional, the reconstructed geometry varies little. \YL{Wu et al.}~\cite{wu2018alive} \YL{reconstruct} 3D stylized faces from 2D \YL{caricature images.} \YL{To address the limited deformation space of 3DMM for 3D caricatures, their approach deforms a 3D standard face by} optimizing deformation gradients under the constraints of facial landmarks. The \YL{follow-up} work \cite{cai2021landmark} utilizes \YL{a} CNN to learn the deformation gradients. These methods suffer from poor reconstruction accuracy due to the sparsity of supervision, and the large gap between the standard face and the target. Following \cite{saito2019pifu}, \YL{Qiu et al.}~\cite{qiu20213dcaricshop} \YL{predict} the surface of 3D caricatures with \YL{an} implicit function,
\YL{which is then aligned with 3DMM}.
However, this method requires \YL{a} large amount of 3D training data, which is 
\YL{difficult} to collect. Overall, research on reconstructing 3D stylized faces 
\YL{is still quite limited, and}
cartoon reconstruction remains to be a challenging task.

\paragraph{Face Rigging.} Face rigging is a crucial \YL{step} for 3D facial animation. By introducing 3DMM, facial \YL{expressions} can be represented by linear \YL{combinations} of PCA \YL{(Principal Component Analysis)} \YL{basis}~\cite{blanz1999morphable,blanz2003reanimating}. \YL{Vlasic et al.}~\cite{vlasic2006face} \YL{propose} a multi-linear model to encode facial identity, expression and viseme. \YL{Synthesizing} from large quantity of real-world data, these PCA models are generally built without semantics, increasing difficulty of \YL{using them to drive} facial animation. To generate user-specific blendshapes for each neutral \YL{face}, hand-crafted or 3D-scanned blendshape models are required 
\cite{alexander2010digital,lewis2014practice}. \YL{Li et al.~}\cite{li2010example} \YL{generate} facial blendshape rigs from sparse exemplars. \YL{However, it} still relies on existing well-crafted face models, \YL{and} preparing examplars for each \YL{subject} is impractical. \YL{Pawaskar et al.}~\cite{pawaskar2013expression} \YL{transfer} a set of facial blendshapes from one identity to another; however, the topological difference between the two models could have negative impact on its performance. \YL{Some other works}~\cite{garrido2016reconstruction,ichim2015dynamic,casas2016rapid} automatically generate personalized blendshapes from video sequences or RGBD frames. Although these works \YL{achieve} impressive performance, they \YL{require temporally} continuous data, \YL{so are} not applicable \YL{to} single image reconstruction. 

\section{Method}

As \YL{shown in} Fig.~\ref{fig:pipeline}, our pipeline can be split into three parts: stylization, reconstruction, and rigging. For stylization, existing methods like StyleGAN~\cite{karras2019style} have \YL{achieved} impressive performance. Therefore, we directly apply a StyleGAN-based style transfer method~\cite{pinkney2020resolution} to generate cartoon images from real-world portraits. In this section, we will focus on our reconstruction and rigging \YL{methods}.

To recover accurate geometry and detailed texture from a single cartoon image, we split the reconstruction into two stages. The first stage is to make a coarse estimation of the face geometry with CNN-based 3DMM coefficients regression. The second stage is to align the face geometry to the input image with fine-grained \YL{Laplacian} deformation. The two-stage reconstruction is designed for cartoon faces with exaggerated shapes by extending the representation space of the low-dimensional 3DMM. 
Finally, to animate the reconstructed model, we transfer the pre-defined expression basis from the standard face to the user-specific face for semantic-preserving facial rig generation. 

\subsection{Model-based Coarse Reconstruction} 

\subsubsection{Template Models: 3DMM}

Expressed by 3D meshes, human faces generally consist \YL{of} a large quantity of vertices and faces to show facial details. During reconstruction, directly predicting each vertex's position is a daunting and time-consuming task. However, human faces share some common geometrical features, such as \YL{the} eyes and \YL{nose}, making it possible to reduce the representation complexity. 3DMM~\cite{blanz1999morphable} was then proposed to encode 3D faces into low-dimensional subspace through linear combinations of shape and texture \YL{bases}:
\begin{align}
    \mathcal{S} &= \overline{\mathcal{S}} + \alpha_{id} A_{id} + \alpha_{exp}A_{exp} \\
    \mathcal{T} &= \overline{\mathcal{T}} + \alpha_{tex} A_{tex}
\end{align}
where $\overline{\mathcal{S}}$ and $\overline{\mathcal{T}}$ represent the shape and texture of the standard face. $A_{id}$, $A_{exp}$, and $A_{tex}$ are 3DMM \YL{bases} for identity, expression, and texture respectively. These \YL{bases} are extracted and synthesized from a large amount of real facial scans. $\alpha_{id}$, $\alpha_{exp}$, and $\alpha_{tex}$ are combination coefficients of the \YL{bases}. $\mathcal{S}$ and $\mathcal{T}$ are the shape and texture of a 3D face. Our model-based reconstruction utilizes 3DMM to make a coarse estimation of the face geometry due to its \YL{expressiveness and} simplicity. 

\subsubsection{Coarse 3D Cartoon Face Reconstruction}

Based on previous CNN-based methods~\cite{guo2018cnn,deng2019accurate}, we utilize a CNN to predict 3DMM coefficients. As Fig.~\ref{fig:recon_pipeline}(a) shows, the network takes a 2D cartoon image as input, \YL{and} predicts a vector of coefficients $x = (\alpha_{id}, \alpha_{exp}, \alpha_{tex}, \delta, p)$. The 3D face pose $p$ \YL{in} the world \YL{coordinate system} is defined \YL{as} a 
\YL{rigid body}
transformation with rotation $R\in \mathrm{SO(3)}$ and translation $t\in \mathbb{R}^3$. $\delta$ is the Sphere Harmonic (SH) coefficients to estimate the global illumination of a \YL{Lambertian} surface on each vertex as $\Phi(n_i,b_i|\delta) = b_i\cdot \sum_{k=1}^{B^2} \delta_k \phi_k (n_i)$, 
where human faces are assumed to be \YL{Lambertian surfaces}~\cite{guo2018cnn,ramamoorthi2001efficient}, $\phi_k:\mathbb{R}^3\rightarrow \mathbb{R}$ represents SH basis functions $(1 \leq k \leq B^2$), and $\Phi(n_i,b_i|\delta)$ computes the irradiation of a vertex with normal $n_i$ and scalar albedo $b_i$. Applying these coefficients to 3DMM \YL{gives the} reconstructed 3D face.

To train the network, 
we first render the face image from the predicted 3D face model at pose $p$ and lighting approximation $\delta$ using differential rendering~\cite{Laine2020diffrast} techniques. The rendered image $\mathrm{I}_{render}$ is then compared with the input image $\mathrm{I}_{in}$ to calculate the loss. 

Specifically, the loss function consists of three parts:
\begin{equation}
    \mathcal{L}(x)=\omega_{l}\mathcal{L}_{lan}(x)+\omega_{p}\mathcal{L}_{photo}(x)+\omega_{r}\mathcal{L}_{reg}(x)
    \label{eq:loss_total}
\end{equation}

The first part is landmark loss:
\begin{equation}
    \mathcal{L}_{lan}(x)=\frac{1}{N}\sum_{n=1}^N\omega_n\|q_n-\Pi(R{\bf p_n}+t)\|^2
    \label{eq:loss_landmark}
\end{equation}
where $q_n\in \mathbb{R}^2$ is the true \YL{position} of the $n$th 2D facial landmark on the original image, ${\bf p_n}\in\mathbb{R}^3$ is the $n$th 3D facial landmark on the face mesh, which is pre-defined by 3DMM. \YL{Note} that 3DMM base models share identical topology, and the related vertices on each base model have the same semantics. 
Therefore, the 3D landmarks could be defined as certain vertices on the mesh. $N$ is the number of landmarks, $\omega_n$ is the loss weight for each landmark, $R$ and $t$ denote the rotation and the transformation of the pose $p$ respectively. \YL{$\Pi = \left[ \begin{array}{ccc} 1 & 0 & 0 \\ 0 & 1 & 0 \end{array} \right]$ is the orthogonal projection matrix from 3D to 2D.}
The second part is photometric loss:
\begin{equation}
    \mathcal{L}_{photo}(x)=\frac{1}{|\mathcal{A}_m|}\|\mathcal{A}_m\cdot(\mathrm{I}_{render}-\mathrm{I}_{in})\|^2
    \label{eq:loss_photo}
\end{equation}
\YL{which} calculates the color difference between $\mathrm{I}_{render}$ and $\mathrm{I}_{in}$ per pixel. $A_m$, acquired by face parsing~\cite{yu2021bisenet}, is the \YL{confidence map} that evaluates whether an image pixel belongs to a human face. This strategy \YL{helps improve} robustness in low-confidence areas, like glasses or beards. Compared to the landmark loss, the photometric loss \YL{constrains} the reconstructed texture and geometry at a fine-grained level. The final part is regularization loss on 3DMM coefficients to avoid getting far from the standard face:
\begin{equation}
    \mathcal{L}_{reg}(x)=\omega_{id}\|\alpha_{id}\|^2+\omega_{exp}\|\alpha_{exp}\|^2+\omega_{tex}\|\alpha_{tex}\|^2
    \label{eq:loss_regularization}
\end{equation}

\subsubsection{Training with Cartoon Data}
\label{sec:cartoon_data}

Most CNN-based methods train their reconstruction network with 
\YL{normal face images}.
However, domain gaps exist between real and cartoon faces. To solve this problem, we propose a cartoon face dataset with landmark labels for network training.

Cartoon face \YL{images} are not as common as real-world \YL{images}. To gather a large amount of cartoon data, we utilize a pre-trained StyleGAN~\cite{karras2019style} for cartoon face generation. Specifically, a StyleGAN is trained on a set of cartoon face images collected from the internet. Then we randomly sample latent codes from the input latent space $\mathcal{Z}$, forward them to the StyleGAN and get the cartoon face \YL{images}. To ensure a clear face appears on each image, we filter out images where face \YL{detection} confidence is lower than a threshold $\epsilon$ using a face detector~\cite{zhang2016joint}.

\begin{figure}[t]
\begin{center}
\includegraphics[width=1.0\linewidth]{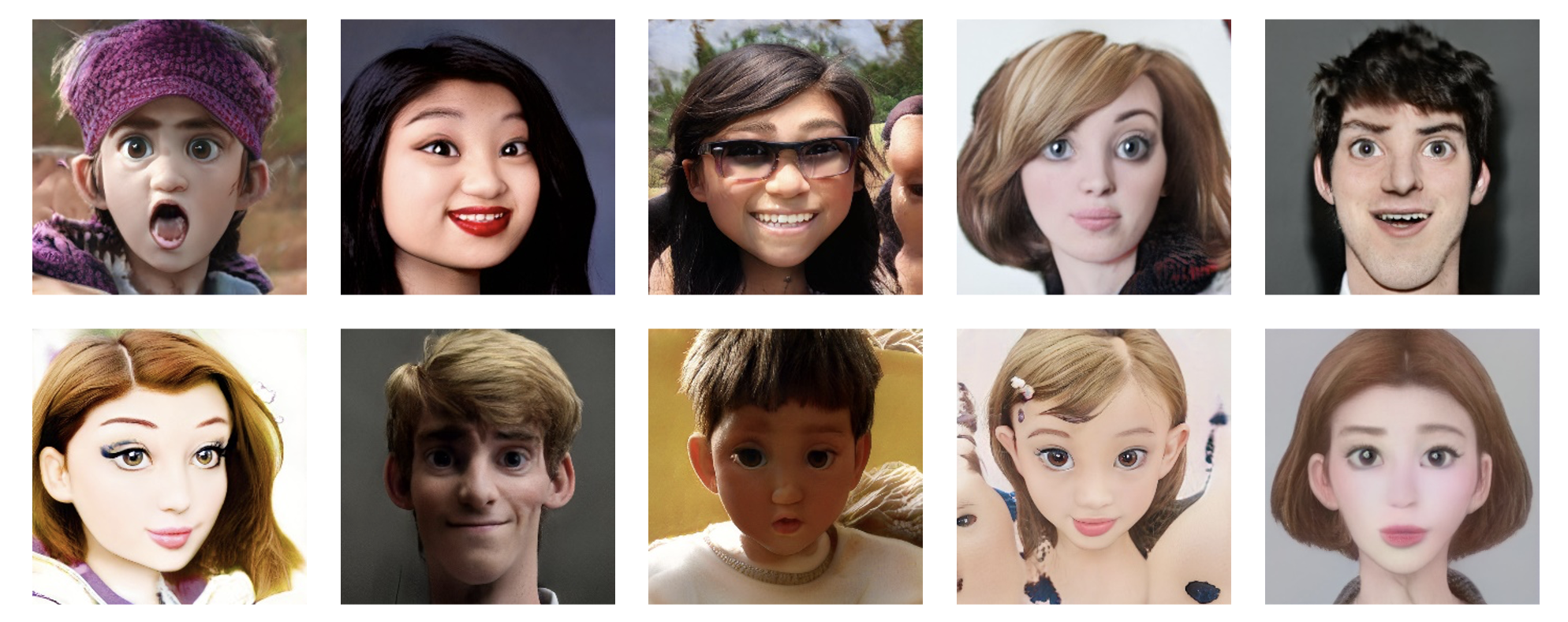}
\end{center}
   \caption{Examples of our cartoon training dataset. For each sample, we ensure that a clear face exists \YL{through a face detector and apply annotations of 68 facial landmarks through a landmark detector}.
   }
\label{fig:cartoon_dataset}
\end{figure}

Fig.~\ref{fig:cartoon_dataset} shows \YL{some} examples of  our cartoon dataset, which contains 73852 images at the resolution of $1024\times1024$. The faces of different colors and ages are uniformly distributed in the dataset to minimize the bias caused by the data distribution. For each image, to calculate the landmark loss (Eq.~\ref{eq:loss_landmark}), \YL{68 landmarks} are labeled by a landmark detector~\cite{king2009dlib}; see section~\ref{sec:2dann} for more details.

In addition, we use the same StyleGAN structure with a ``layer swapping'' interpolation scheme~\cite{pinkney2020resolution} to stylize users' real-world portraits. These images then become the input of the coarse reconstruction process in our complete application pipeline. The size of the stylized image is currently fixed in this work. However, works of image enhancement~\cite{zhang2021self,muslim2019knowledge} have shown the potential to increase the size and resolution of the image. Thus the image size won't be a limitation of this work.

\subsection{Deformation-based Fine Reconstruction}

Although using 3DMM for coarse reconstruction yields accurate results on the overall shape of the face, we find it fails to recover some fine face structures, especially the eyes. The low-dimensional parametric face model lacks expressivity for exaggerated facial parts, which are common in cartoon portraits. These reconstruction errors \YL{cannot} be ignored because even a tiny misalignment would significantly affect the model appearance and facial animation.

To tackle this issue, we introduce deformation-based fine reconstruction. As Fig.~\ref{fig:recon_pipeline}(b) shows, we align the 3D reconstructed face to the 2D landmarks on the input image with non-rigid deformation. We minimize the misalignment with accurate landmark supervision and \YL{a} local deformation method. We show that our facial alignment strategy significantly improves texture mapping performance.

\subsubsection{Cartoon Face 2D Landmark Annotation} \label{sec:2dann}

Accurate 2D landmark annotation is crucial to the alignment. We observe that significant misalignment appears in the eye areas after projecting the predicted 3D face to the image space. Some mainstream 68-landmark detectors~\cite{king2009dlib}, which are trained on \YL{ordinary face images}, could provide landmark annotations on the image. However, the annotation is not accurate on cartoon \YL{images}, especially \YL{in the eye} \YL{areas}, because of the domain gap. To solve this problem, we combine landmark detection with a state-of-the-art pixel-level face parsing method~\cite{yu2021bisenet}. We first \YL{obtain the prediction of 68 facial landmarks} from the detectors and acquire \YL{the} face parsing result, which contains \YL{eye} segmentation. Then, for each eye landmark, we \YL{snap} its position to the nearest point on the boundary of the segmented eye area if the boundary exists. Utilizing color clues, we set the eye landmarks \YL{to lie} on the border of the eye.

\subsubsection{Face Alignment with Laplacian Deformation} 

An intuitive way to align \YL{a} 3D face with 2D landmark labels is to optimize the 3DMM coefficients by minimizing the distance between the projected 3D landmarks and the 2D labels:
\begin{align}
    \alpha_{id}^*, \alpha_{exp}^* &= \underset{\alpha_{id}, \alpha_{exp}}{\arg\min} \sum_{n=\YL{1}}^N
    \omega_n\|q_n-\Pi(R {\bf p_n}+t)\| \label{eq:template_refine}\\
    {\bf p_n} &= K(\overline{\mathcal{S}} + \alpha_{id} A_{id} + \alpha_{exp}A_{exp}; n)
\end{align}
where $q_n$, ${\bf p_n}$, $\Pi$, and $(R, t)$ share the same definition as Eq.~\ref{eq:loss_landmark}. $K(\mathcal{S};n)\in\mathbb{R}^3$ is to get the $n$th 3D landmark position on shape $\mathcal{S}$. However, 
adjusting 3DMM coefficients in this way will cause distortion and unnatural folds on the face due to \YL{the global nature and} geometric restrictions of the template models, which will be demonstrated in the experiments (section \ref{sec:align_exp}).

We \YL{instead} exploit \YL{Laplacian} deformation~\cite{zhou2005large} to align the landmarks accurately and locally without affecting the overall shape. The deformation is driven by anchors, which are landmarks in this context. The goal is to preserve the local normal of each vertex on the mesh as much as possible while moving the anchors. Specifically, the \YL{Laplacian coordinates} of vertex ${\bf v_i}$ \YL{are} defined as:
\begin{equation}
    L({\bf v_i}) = \frac{1}{|\YL{\mathcal{N}}({\bf v_i})|}\sum_{{\bf v_j}\in \YL{\mathcal{N}}({\bf v_i})}({\bf v_i}-{\bf v_j})
    \label{eq:laplacian_coord}
\end{equation}
where $\YL{\mathcal{N}}({\bf v_i})$ is the set of vertices that share common edges with ${\bf v_i}$ \YL{(i.e., 1-ring neighboring vertices)}. Preserving $L({\bf v_i})$ during deformation imposes a constraint on local geometry that prevents unnatural distortions. Meanwhile, to be driven by anchors, corresponding vertices should follow the anchors and stay close. Therefore, the objective function to be minimized is:
\begin{equation}
    \min_{{\bf v}\in V} (\sum_{i=1}^{|V|}\|L({\bf v_i})-L_i^\prime\|^2+\lambda\sum_{k\in M}\|{\bf v_k}-{\bf p_k}\|^2)
    \label{eq:laplacian_optim}
\end{equation}
where $L_i^\prime$ is \YL{the} initial \YL{value} of $L({\bf v_i})$, \YL{$M$ is the set of vertex indices for 3D landmarks on the mesh, and}
${\bf v_k}\in\mathbb{R}^3$ \YL{is a 3D landmark position,} and ${\bf p_k}\in\mathbb{R}^3$ \YL{is the corresponding ground truth 3D position}. 
Transforming 2D landmark supervision $q_n$ to 3D anchors ${\bf p_k}$ requires depth information. We use the depth value of the initial 3D landmark vertex ${\bf v_k}$ as an approximation of ${\bf p_k}$'s:
\begin{equation}
    d_{cam}-(R{\bf p_k}+t)\big{|}_z=d_{cam}-(R{\bf v_k}+t)\big{|}_z
\end{equation}
where $d_{cam}$ is the depth of the camera center, $(R, t)$ is a \YL{rigid body} transformation to the camera coordinate \YL{system}.

\subsubsection{Texture Mapping}

\begin{figure}[t]
\begin{center}
\includegraphics[width=1.0\linewidth]{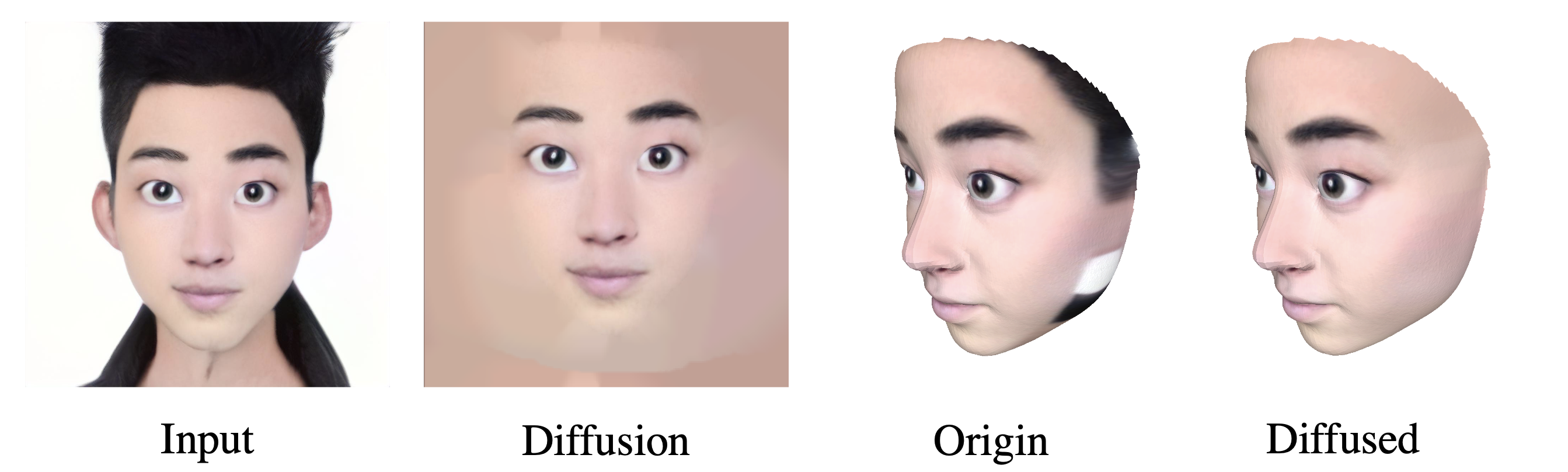}
\end{center}
   \caption{Texture diffusion (left 2). Comparison of the \YL{results} without (Origin) and with (Diffused) diffused texture (right 2).}
\label{fig:diffused_texture}
\end{figure}

Texture plays a decisive role in improving the visual quality of the reconstructed model. The texture we acquired from coarse reconstruction is \YL{a} combination of 3DMM \YL{texture basis}, which is too rough to express an elaborate cartoon face. Therefore, to maximize the similarity of the model with the input cartoon image, we project each vertex to the image with the transformation $(R, t)$ \YL{predicted} in the coarse reconstruction stage. The normalized 2D projected position \YL{is then used as} the texture \YL{coordinates} of the vertex:
\begin{equation}
    {\rm tex\_coord}({\bf v}) = {\rm Norm}(\Pi(R {\bf v}+t))
\end{equation}

\paragraph{Diffused Texture.} Due to tiny reconstruction inaccuracy, some background pixels might be mistakenly mapped as part of the face texture. This error will be amplified on the 3D model, as Fig.~\ref{fig:diffused_texture} (Origin) shows. To tackle this problem, we first segment the cartoon face from the background with face parsing~\cite{yu2021bisenet}, \YL{and} then replace the background with diffusion of the face color, as Fig.~\ref{fig:diffused_texture} (Diffusion) shows. Each background pixel is traversed by a Breadth-First-Search, and \YL{its} color is replaced with the average color of the surrounding visited pixels. The processed image is then used for texture mapping.

\subsection{Semantic-preserving Facial Rig Generation}

Animating a static 3D cartoon face requires additional action guidance. Motivated by 3DMM, we utilize a template-based method for facial animation:
\begin{equation}
   S^*=S_0+B_{exp}\beta
   \label{eq:temp_anim}
\end{equation}
where $S_0$ is the neutral 3D face, \YL{and} $B_{exp}$ is the expression \YL{basis}. Controlled by coefficients $\beta$, the output face $S^*$ changes expression \YL{accordingly}. Normally, the expression \YL{components} of 3DMM \YL{basis} lack semantics and are mutually coupled, making it difficult to control each part of the face independently. Inspired by FACS~\cite{ekman1978facial}, we manually construct a set of standard face models $\{S_i\}, i=1,2,3...,m$, each of which represents a \YL{specific} movement of a single face part, such as `left eye close' and `mouth open.' Then we have $B_{exp}=(S_1-S_0, S_2-S_0,...,S_m-S_0), \beta=(\beta_1, \beta_2,...,\beta_m)$. Generally, $\beta_i$ ranges from 0 to 1.

\begin{figure}[t]
\begin{center}
\includegraphics[width=1.0\linewidth]{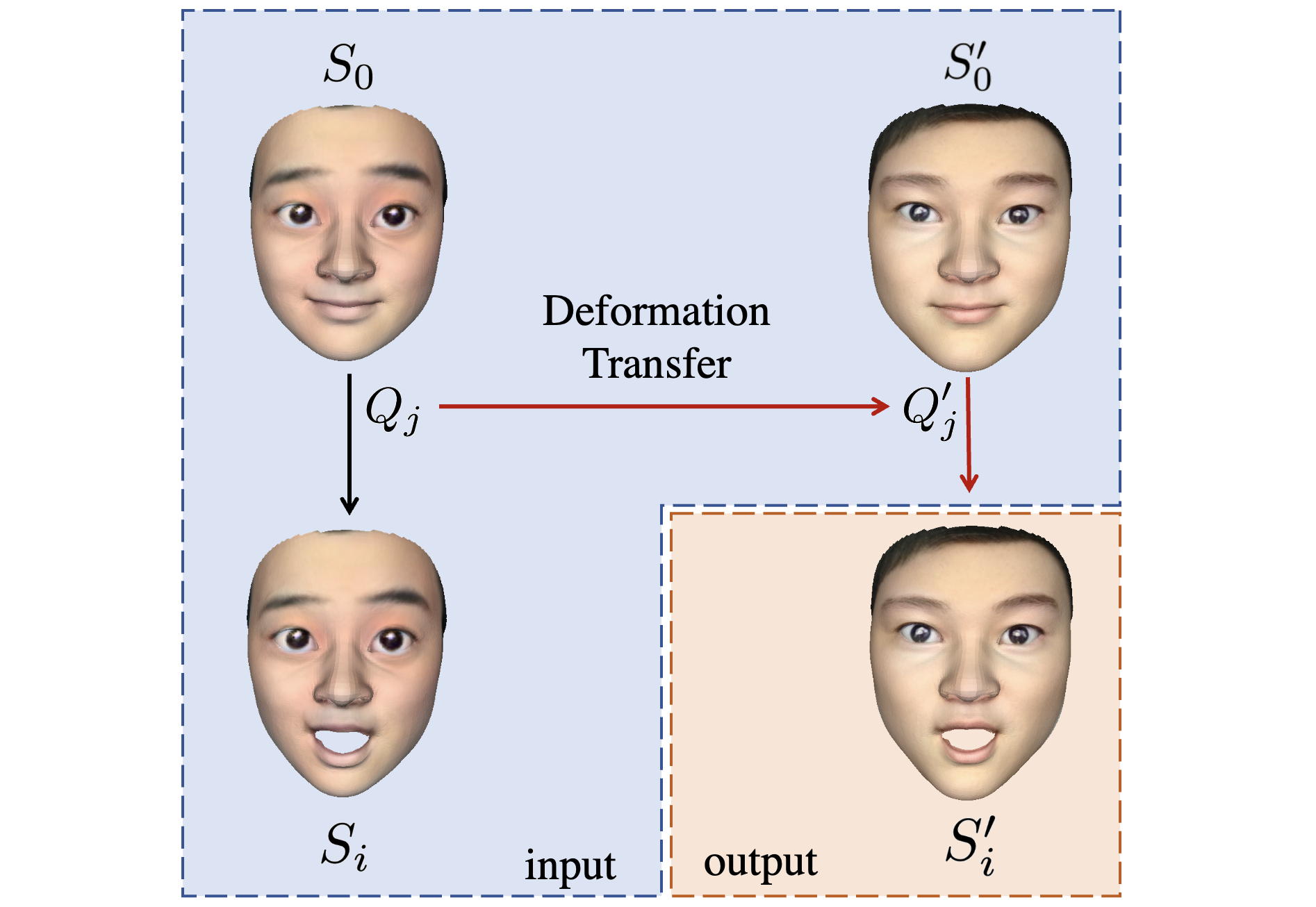}
\end{center}
   \caption{Expression Transfer. The deformation from $S_0$ to $S_i$ (\YL{$Q_j$ for face $f_j$}) is transferred to the deformation from $S_0^\prime$ to $S_i^\prime$ (\YL{$Q_j^\prime$ for face $f_j$}). By generating new expression models $S_i^\prime$ \YL{using the subject-specific expression models,}  $S_0^\prime$ can be animated.}
\label{fig:expression_trans}
\end{figure}

However, directly applying the standard expression models $\{S_i\}$ to an arbitrary neutral face will cause unnatural expressions, \YL{due to differences of facial shapes}. \YL{So we} utilize deformation transfer~\cite{sumner2004deformation} to generate user-specific face rig. As Fig.~\ref{fig:expression_trans} shows, the deformation from $S_0$ to $S_i$ is transferred to adapt \YL{to} the newly reconstructed $S_0^\prime$ and generate $S_i^\prime$. The expression transfer is based on the geometric relations between the standard neutral face $S_0$, $S_i$ and $S_0^\prime$. 

As to the deformation from $S_0$ to $S_i$, since they are topologically \YL{consistent}, vertices and faces between them correspond \YL{to each other}. 
For a triangular face \YL{$f_j$}, suppose ${\bf v_i}$ and $\Tilde{{\bf v_i}}$, $i=1,2,3$, are undeformed and deformed vertices of \YL{$f_j$} respectively. To include normal information, \cite{sumner2004deformation} introduces the fourth vertex ${\bf v_4}$ in the direction perpendicular to \YL{$f_j$} \YL{with a unit distance} as:
\begin{equation}
    {\bf v_4} = {\bf v_1}+\frac{({\bf v_2}-{\bf v_1})\times({\bf v_3}-{\bf v_1})}{\sqrt{|({\bf v_2}-{\bf v_1})\times({\bf v_3}-{\bf v_1})|}}
\end{equation}
Then the deformation of $f_j$ \YL{can} be described with a $3\times 3$ matrix $Q_j$ and a translation vector $t_j$ as:
\begin{equation}
    \Tilde{{\bf v_i}} = \YL{Q_j}{\bf v_i}+\YL{t_j}, i=1,2,3,4
\end{equation}
As to the transformation from $S_i$ to $S_i^\prime$, the goal is to preserve $Q_j$'s:
\begin{equation}
    \min_{\Tilde{\bf v}_1^\prime, ..., \Tilde{\bf v}_n^\prime} \sum_{j=1}^m\|Q_j-Q_j^\prime\|
    \label{eq:exp_trans}
\end{equation}
where $Q_j$ is the transformation matrix of the $j$th triangular face on mesh from $S_0$ to $S_i$, $Q_j^\prime$ is for $S_0^\prime$ to $S_i^\prime$; $m$ is the number of faces and $\{\Tilde{\bf v}_1^\prime, ..., \Tilde{\bf v}_n^\prime\}$ are vertices of $S_i^\prime$. \YL{$t_j$ remains unchanged when transferred to $S_0\prime$.}

We now can obtain the expression models $\{S_i^\prime\}$ for the newly reconstructed model by applying the above expression transfer to each $\{S_i\}$. Then the 3D face can be animated in real-time driven by the input coefficients $\beta$.

\section{Experiments}

\subsection{Setup}

\paragraph{Implementation Details.} We implement the coarse reconstruction network \YL{using} the \YL{PyTorch} framework~\cite{paszke2019pytorch}. The network takes a stylized face image with size $224\times224\times3$ as input, and outputs a coefficient vector $x\in \mathbb{R}^{239}$, with $\alpha_{id}\in\mathbb{R}^{80}$, $\alpha_{exp}\in\mathbb{R}^{64}$, $\alpha_{tex}\in\mathbb{R}^{9}$, $\delta\in\mathbb{R}^{6}$ respectively. In our experiment, we set the weights to $\omega_{id}=1.2$, $\omega_{exp}=1.0$, $\omega_{tex}=1.2e-3$, $\omega_l = 2e-3$, $\omega_p=2.0$, $\omega_r = 3e-4$. Similar to \cite{deng2019accurate}, we use a ResNet-50 network as the backbone followed by a fully-connected layer to regress the coefficients. For the fine reconstruction stage, the optimization problem \YL{in} Eq.~\ref{eq:laplacian_optim} can be transformed into a linear equation by the least \YL{squares} method. We solve the linear equation with sparse \YL{matrices} and Cholesky decomposition. The same processing is \YL{applied} to the expression transfer optimization problem \YL{in} Eq.~\ref{eq:exp_trans} in facial rig generation. Our manually constructed standard expression models are built on blender~\cite{blender} by professional modelers, containing 46 different expressions defined by FACS~\cite{ekman1978facial}.

\paragraph{Data Collection.} As \YL{introduced in section~}\ref{sec:cartoon_data}, we built a training dataset with 73852 cartoon \YL{face} images for coarse reconstruction training. For testing data, we collect real-world portraits and stylize them using a pretrained StyleGAN~\cite{karras2019style}. We then annotate 68 facial landmarks for each stylized cartoon image with the landmark detector\cite{king2009dlib} and manually adjust their positions. The \YL{test} set contains 50 images with various lighting conditions and shapes.

\subsection{Results on Cartoon Face Reconstruction}

\subsubsection{Comparison with Prior Art}

\begin{table*}
\begin{center}
\begin{tabular}{lcccccccc}
\toprule
\multirow{2}{*}{Method} & \multicolumn{6}{c}{Landmark Error$\downarrow$} && \multirow{2}{*}{Photometric Error$\downarrow$} \\
\cmidrule{2-7}
 & eyes & nose & brow & mouth & contour & total &&\\
\midrule
PRN~\cite{feng2018joint}  & 269.45 & 200.41 & 156.34 & 397.67 & 435.86 & 1459.73 && 1.06\\
Deep3D~\cite{deng2019accurate} & 100.23 & 0.51 & 11.60 & 1.57 & 16.17 & 130.08 && 3.31\\
CoarseRecon (Ours) & 98.33 & 0.55 & 11.40 & 1.54 & 16.14 & 127.96 && 3.27\\
FineRecon (Ours) & \textbf{8.27} & \textbf{0.23} & \textbf{11.33} & \textbf{1.50} & \textbf{16.11} & \textbf{37.44} && \textbf{0.83} \\
\bottomrule
\end{tabular}
\end{center}
\caption{\YL{Comparison} with \YL{prior arts}.}
\label{tab: comparison}
\end{table*}

We compare our method with PRN~\cite{feng2018joint}, a template-free method that predicts face shapes with a CNN, and Deep3D~\cite{deng2019accurate}, a baseline that predicts 3DMM coefficients in an unsupervised manner. Both works have been proposed recently, showing impressive performance on 3D face reconstruction. We also report the results of our two stages: coarse reconstruction and fine reconstruction, to validate the effectiveness of the two-stage design. We measure the reconstruction quality by computing the 2D landmark and photometric errors on the \YL{test} set. Specifically, for each \YL{test} image, we project the result to the image plane after reconstruction. The landmark error \YL{measures} the \YL{Euclidean} \YL{distances} between the projected landmarks and the annotations, evaluating the correspondence and shape accuracy. We evaluate the error of different face parts separately. We also use the photometric error, which is the average \YL{Manhattan} distance of the pixel colors between the rendered image and the input image, to evaluate the appearance similarity. We show the average results over the \YL{test} data.

\begin{figure}[t]
\begin{center}
\includegraphics[width=1.0\linewidth]{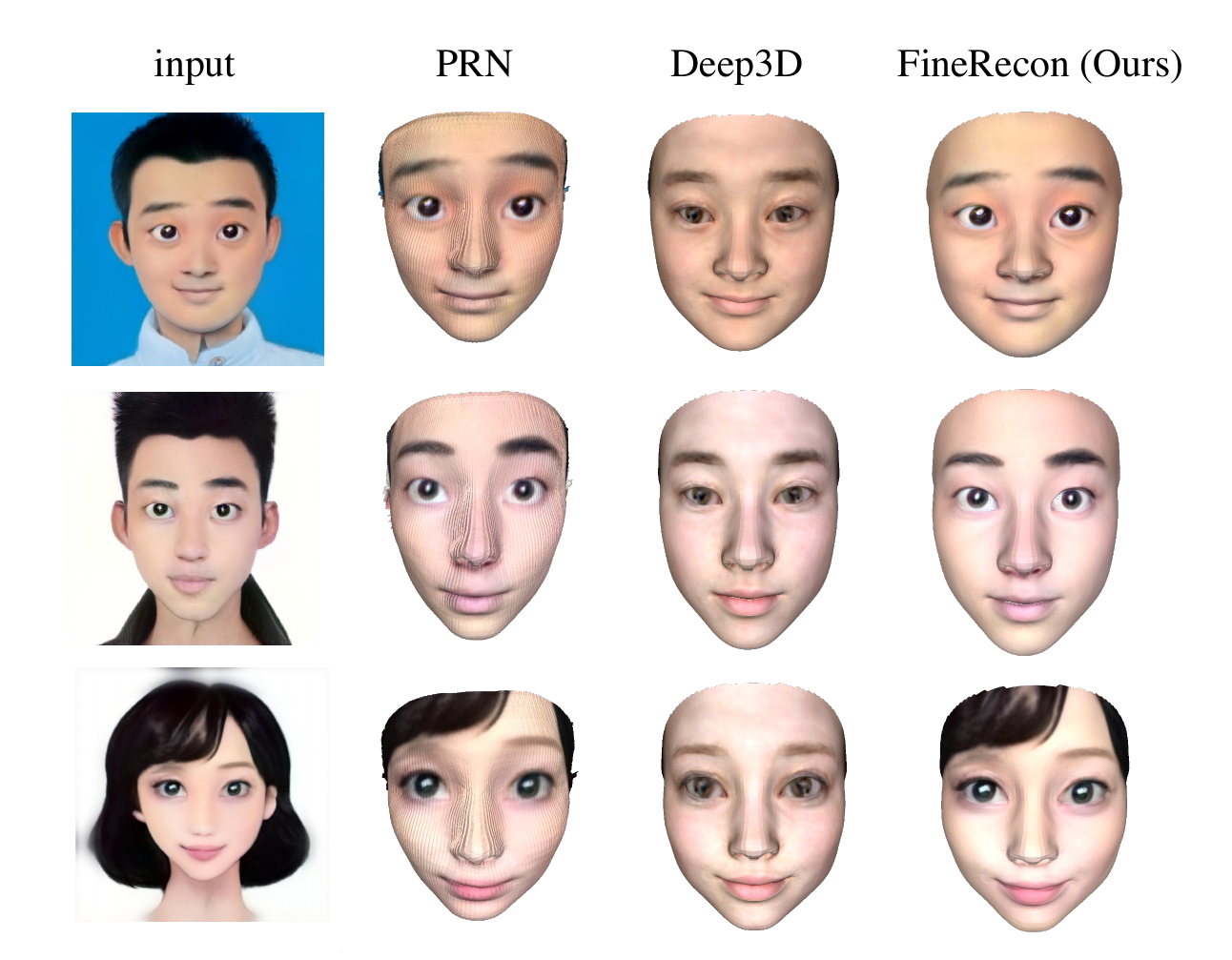}
\end{center}
   \caption{Comparison of our results with PRN and \YL{Deep3D}.}
\label{fig:comparison}
\end{figure}

As Table~\ref{tab: comparison} shows, our method achieves a much lower landmark error than PRN and \YL{Deep3D}. Although sharing a similar network structure, our coarse reconstruction slightly outperforms \YL{Deep3D} due to the cartoon data training.
Compared with coarse reconstruction, \YL{our} fine reconstruction significantly improves the eyes' alignment accuracy. The accuracy of other facial parts like the nose, eyebrow, and mouth are also improved. It validates the effectiveness of our deformation-based alignment strategy. To map texture from the input image, alignment with the image should be accurate. Otherwise, it would cause evident unnatural facial colors. Our fine reconstruction also achieves the lowest photometric error due to accurate reconstruction, alignment, and texture mapping. Although PRN utilizes the input image for texture mapping like our method, which is the reason why PRN's result looks similar to the input image, it has a larger photometric error because the inaccuracy of the shape and alignment causes background pixels to be mistakenly mapped to the texture. We show visualization comparisons in Fig.~\ref{fig:comparison}.


\subsubsection{Evaluation on Face Alignment}
\label{sec:align_exp}

\begin{table}
\begin{center}
\begin{tabular}{lccc}
\toprule
& Adjust Exp& Adjust Id+Exp& Ours\\
\midrule
Landmark Error$\downarrow$& 143.38& 110.70& \textbf{37.44} \\
\bottomrule
\end{tabular}
\end{center}
\caption{Comparison with template-based method.}
\label{tab: alignment}
\end{table}

\begin{figure}[t]
\begin{center}
\includegraphics[width=1.0\linewidth]{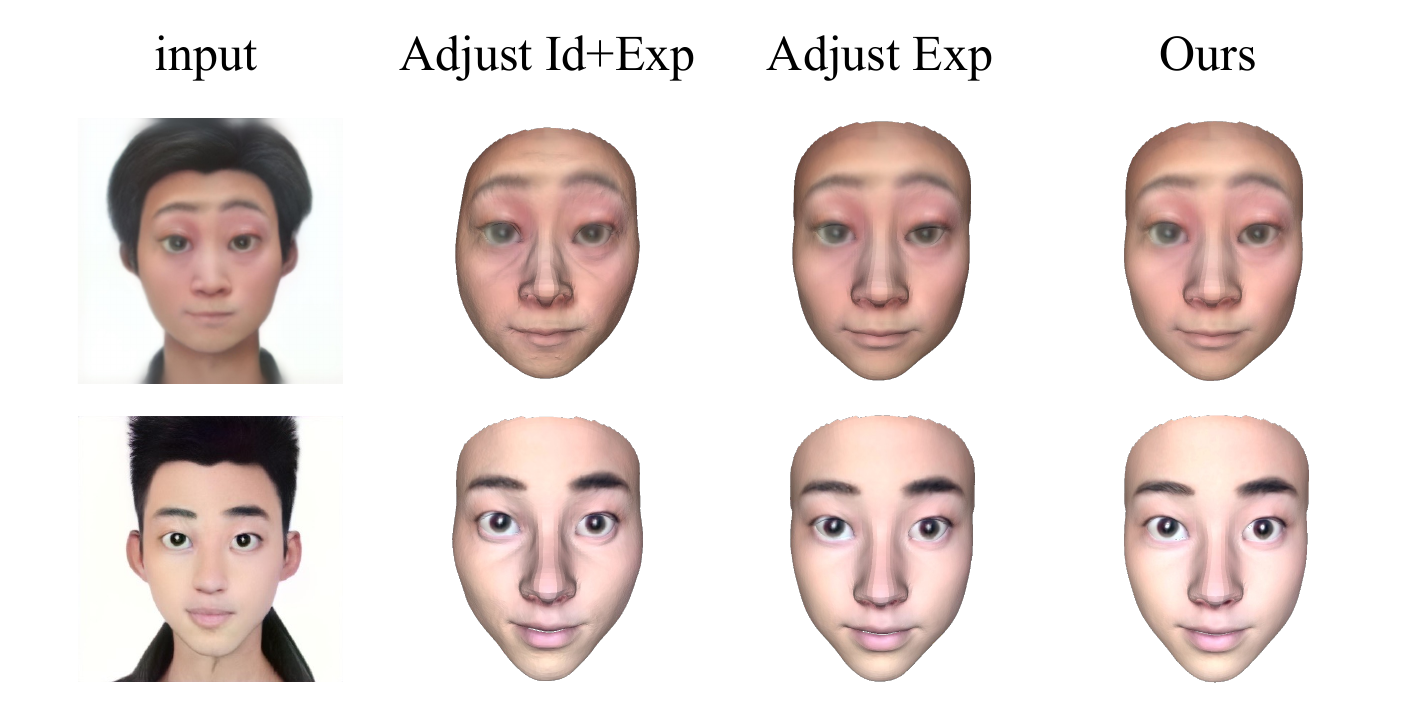}
\end{center}
   \caption{Comparison of the results using different face alignment \YL{strategies}: template-based method (Adjust Id+Exp/Adjust Exp) and deformation-based method (ours).}
\label{fig:refine}
\end{figure}

\paragraph{Comparison with the template-based method.} Eq.~\ref{eq:template_refine} shows an intuitive way of adjusting 3DMM coefficients $\alpha_{id}$ and $\alpha_{exp}$ to align with 2D landmark labels. There are two schemes to optimize the coefficients based on templates: adjusting $\alpha_{exp}$ only (Adjust Exp), and adjusting both $\alpha_{id}$ and $\alpha_{exp}$ at the same time (Adjust Id+Exp). Table~\ref{tab: alignment} shows a comparison of our deformation-based method with these two template-based methods. We use landmark \YL{errors as the} criteria with the same definition in Table~\ref{tab: comparison}. Our method has lower landmark \YL{errors} than these baselines. Interestingly, `Adjust Id+Exp' gains lower landmark error than `Adjust Exp', due to a higher degree of freedom (DoF) and larger representation space. In this regard, our method has the highest DoF and shows the lowest error. We also demonstrate with visualization \YL{results} in Fig.~\ref{fig:refine}. Although `Adjust Id+Exp' gains lower landmark error than `Adjust Exp', the visualization shows unnatural wrinkles and distortion due to the restrictions of templates. This suggests that the refinement exceeds the template's representation capability. On the other hand, our method can retain high-quality visual performance, \YL{while making} accurate adjustments simultaneously.

\begin{figure}[t]
\begin{center}
\includegraphics[width=1.0\linewidth]{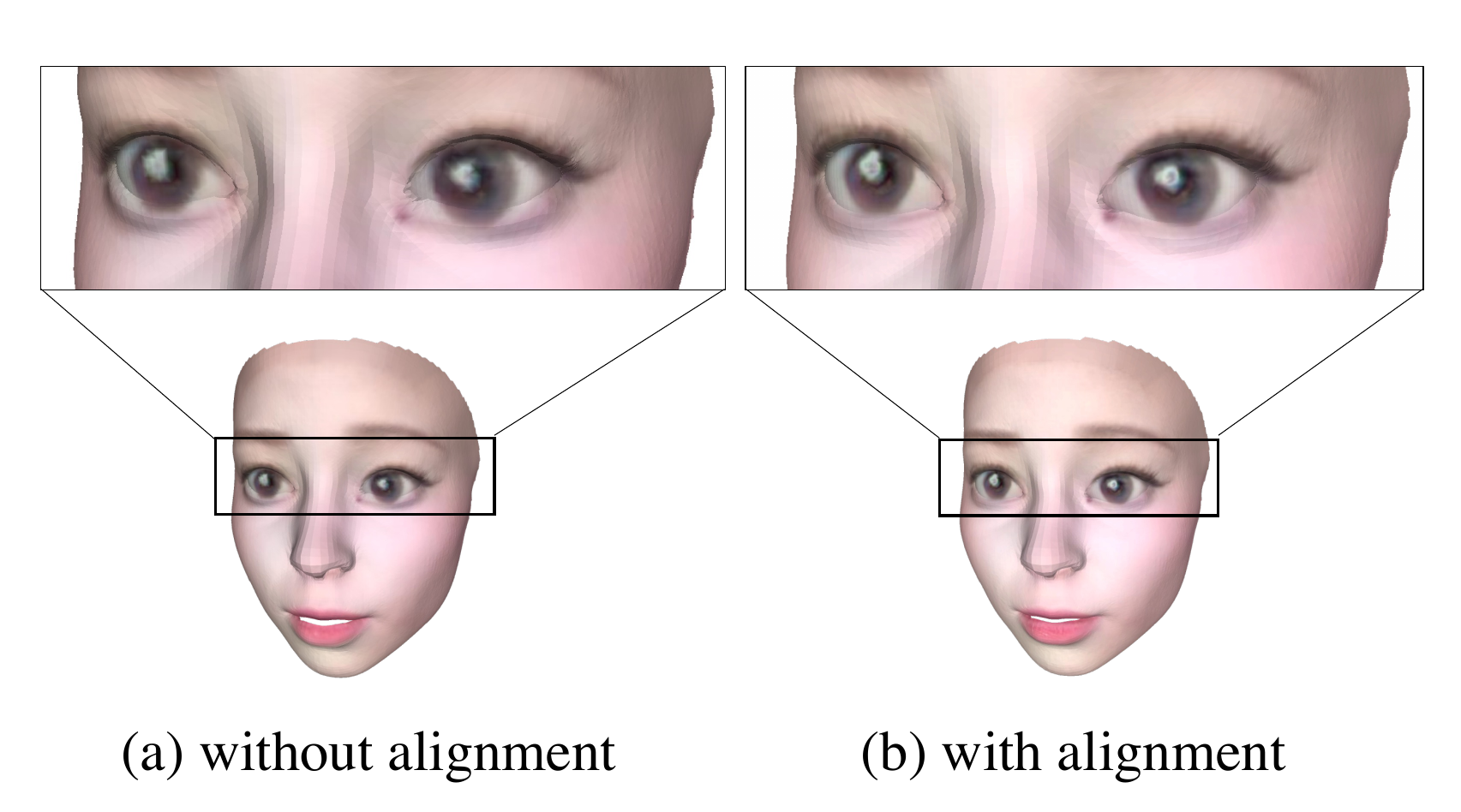}
\end{center}
   \caption{Comparison of the results on \YL{eye areas} without (a) and with (b) face alignment.}
\label{fig:compare_align}
\end{figure}

Fig.~\ref{fig:compare_align} shows the comparison of the results on \YL{eye areas} with or without face alignment. It \YL{is} clear that before the alignment, part of the eye texture is mistakenly mapped to the face skin, because the coarse reconstructed eyes are too small. During animation, the wrongly mapped texture will be amplified, for example, when the eyes are closing.  

\subsubsection{User Subjective Evaluation}

\begin{table}
\begin{center}
\begin{tabular}{lccc}
\toprule
& Aesthetics& Accuracy& Similarity\\
\midrule
PRN~\cite{feng2018joint} & 2.63/1.07 & 2.99/1.04 & 3.12/1.17 \\
Deep3D~\cite{deng2019accurate} & 2.66/1.22 & 2.62/1.01 & 2.46/1.02 \\
Ours & \textbf{3.75/0.92} & \textbf{3.88/0.81} & \textbf{3.95/0.87} \\
\bottomrule
\end{tabular}
\end{center}
\caption{User subjective evaluations. The table shows mean and standard deviation (mean/\YL{std. dev.}) of users' evaluation scores. Our method achieves the highest ratings on \YL{all} three subjective criteria (aesthetics, accuracy and similarity).}
\label{tab: user_eval}
\end{table}

To make a more comprehensive evaluation of our reconstruction results, we conduct \YL{a} user study to collect subjective evaluations of the reconstruction. 
For each participant, we send out a questionnaire with six independent questions. For each question, we randomly select a cartoon face image from the \YL{test} set, \YL{and} reconstruct its 3D model with PRN, \YL{Deep3D} and our method. We show the results of these methods in \YL{a} random order and ask participants to rate for aesthetics, accuracy and similarity. Aesthetics evaluates whether the 3D model is aesthetically pleasing. Accuracy evaluates the correctness of the overall shape and the \YL{position} of each face part. Similarity evaluates whether the 3D model appears similar to the input image. Participants \YL{are asked to rate} \YL{in the range of} 1-5 for each aspect, \YL{where} 1 for very poor, and 5 for very good.

We invited 55 participants in total, 30 males and 25 females distributed from diverse backgrounds. 
Table~\ref{tab: user_eval} shows the average score \YL{and standard deviation} \YL{on all participants for each question and method.} Our method evidently outperforms the other two methods on subjective criteria, including aesthetics, accuracy and similarity. To our observation, face shape and texture play important roles in improving performance on these subjective criteria. The results suggest that 
\YL{our method does not overfit the landmark constraints, but rather it uses the appropriate constraints to achieve overall 
 high visual quality.}

\subsection{Results on Face Rigging and Animation}

\begin{figure}[t]
\begin{center}
\includegraphics[width=1.0\linewidth]{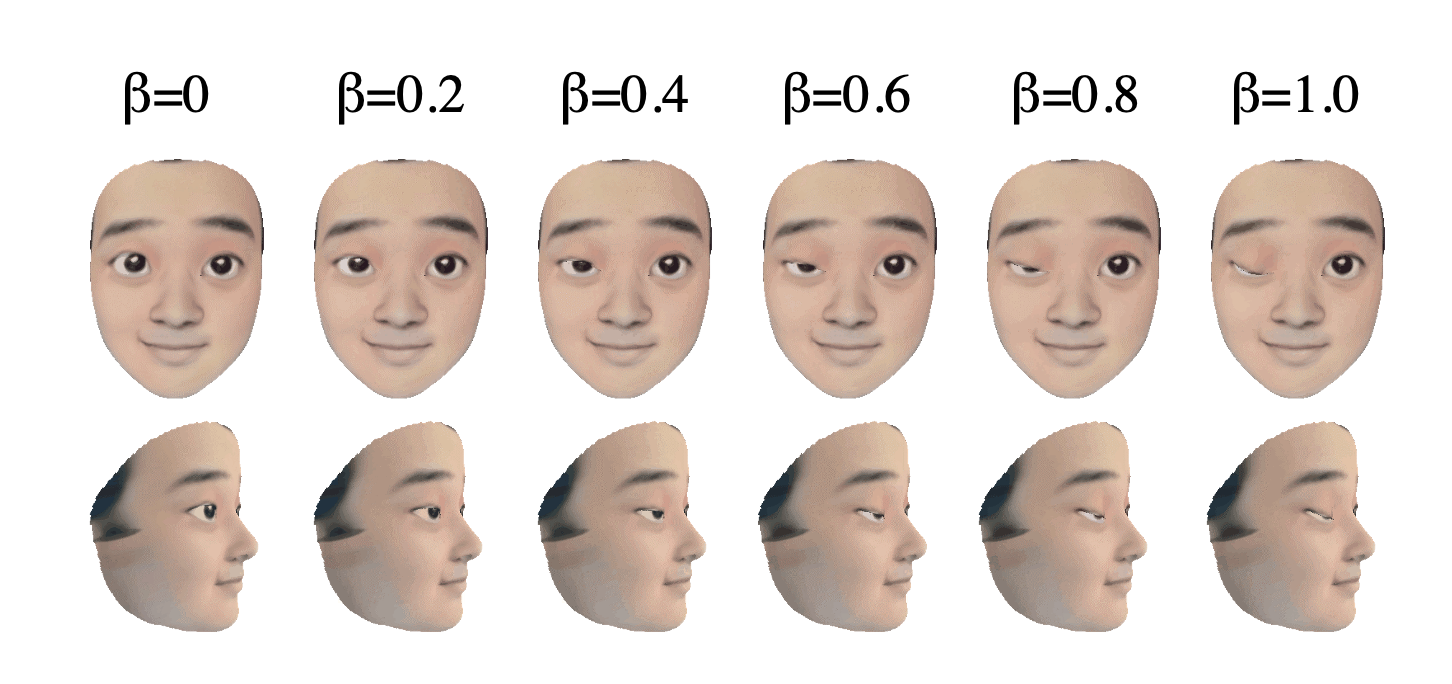}
\end{center}
   \caption{Template-based facial animation.}
\label{fig:animation}
\end{figure}

\paragraph{Visualizations on Template-based Facial Animation.} Fig.~\ref{fig:animation} shows the linear combination of the neutral face $S_0$ and \YL{an} expression template model $S_i$ with coefficient $\beta$, according to Eq.~\ref{eq:temp_anim}.
The semantic of $S_i$ is `right eye close', which allows us to control the right eye independently. We have 46 template models with different semantics such as mouth open, left brows up, lip funnel, etc. 

\begin{figure}[t]
\begin{center}
\includegraphics[width=1.0\linewidth]{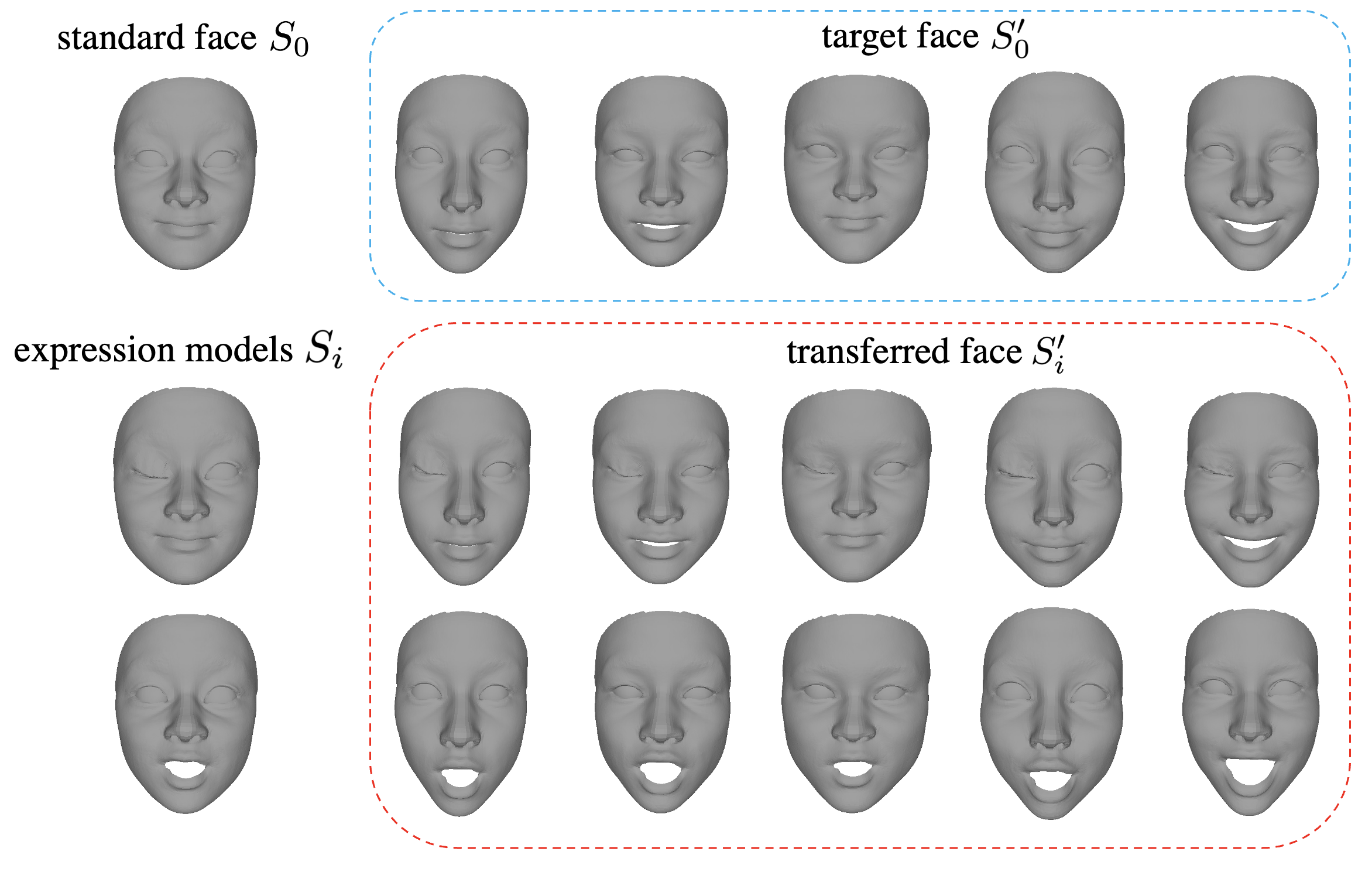}
\end{center}
   \caption{Visualization results of expression transfer on part of the \YL{test} data. We demonstrate two expressions: `mouth open' and `close right eye'.}
\label{fig:exp_transfer}
\end{figure}

\paragraph{Results on Expression Transfer.} We demonstrate the effectiveness of our expression transfer method in Fig.~\ref{fig:exp_transfer}. We hide the texture to show the geometry of the faces clearly. We demonstrate the transfer of two typical expressions: `right eye close' and `mouth open'. The \YL{results show} that, although the target model $S_0^\prime$ varies in shape, the transferred expression can adapt pretty well. This is because, instead of simply applying the vertex shift to the target model, we transfer the transformation of the triangular faces on the mesh.

\paragraph{Eye-ball Modeling.} To animate the eyes without eyeball distortion, we model the eyeballs independently during face rigging. A sphere fits the eyeball area, and then we move the sphere inside the head for a small distance \YL{$\Delta$}
to avoid collision with the eyelids. The texture is correspondingly mapped to the sphere, and the invisible parts are set to white by default. 

\subsection{Application Results}

\paragraph{Efficiency Evaluation.} Generally, applications require high efficiency of reconstruction and animation. \YL{Our experiment is carried out} on \YL{an computer with} an Intel(R) Xeon(R) E5-2678 v3 @ 2.50GHz CPU and a TITAN RTX GPU. We repeat ten times on each test sample and show the average time consumption. As Table~\ref{tab: efficiency} shows, our method takes 24 seconds on average for reconstruction and face rigging, which is acceptable for a user to wait. Currently, the fine reconstruction algorithm is implemented on CPU, and we believe that the efficiency will be largely improved if \YL{this step is sped} up by GPU. For the run-time, results show that our reconstructed model can change its expression with a real-time performance of over 280 FPS.

\begin{table}
\begin{center}
\begin{tabular}{cccc}
\toprule
\multicolumn{3}{c}{Reconstruction} & Run-time\\
\cmidrule(lr){1-3} \cmidrule(lr){4-4}
CoarseRecon & FineRecon & FaceRigging & Deformation\\
\midrule
1.46s & 20.02s & 1.84s & 3.56ms \\
\bottomrule
\end{tabular}
\end{center}
\caption{Efficiency Evaluation. We show that our pipeline can reconstruct an arbitrary face model within 30s, and perform real-time facial animation over 280 FPS.}
\label{tab: efficiency}
\end{table}

\begin{figure}[t]
\begin{center}
\includegraphics[width=1.0\linewidth]{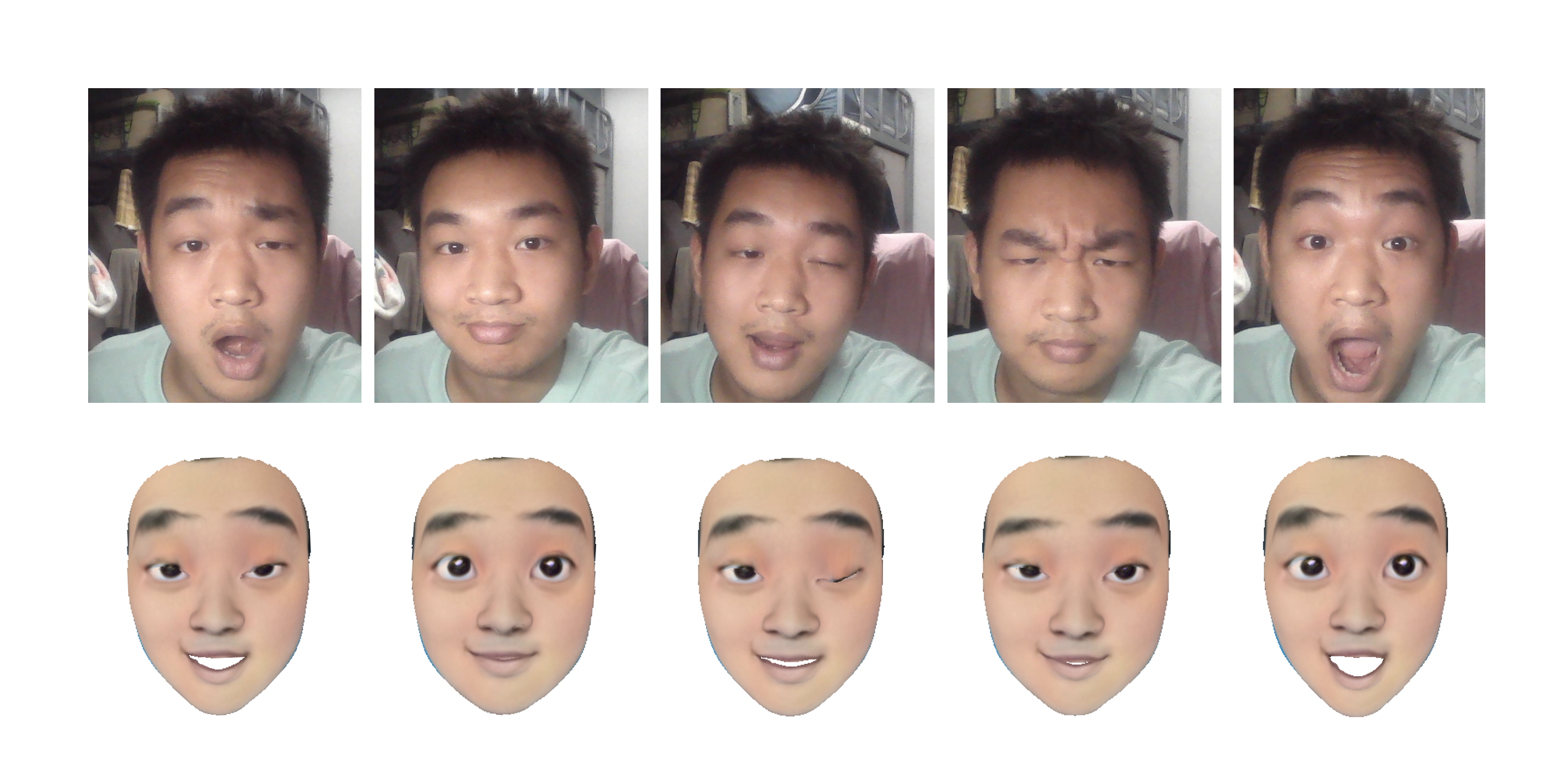}
\end{center}
   \caption{Visualizations of real-time \YL{face-to-face} animation. Our reconstructed model can be driven by a real-world reference face, utilizing an upstream expression driver. }
\label{fig:driven}
\end{figure}

\paragraph{Real-time \YL{Video Driven Face} Animation.} Utilizing a fast expression animation driver~\cite{lugaresi2019mediapipe}, 
we show the potential of real-time \YL{video driven} facial animation in Fig.~\ref{fig:driven}. The upstream driver predicts expression coefficients $\beta$ from a real human face. A reconstructed 3D cartoon face is then animated by $\beta$. The driving process can be implemented online with a separate frontend and backend, where the driver serves as the backend, and the animatable 3D model serves as the frontend. Intuitively, with this functionality, users can drive their own avatars to follow their facial actions in a VR application.

\begin{figure}[t]
\begin{center}
\includegraphics[width=1.0\linewidth]{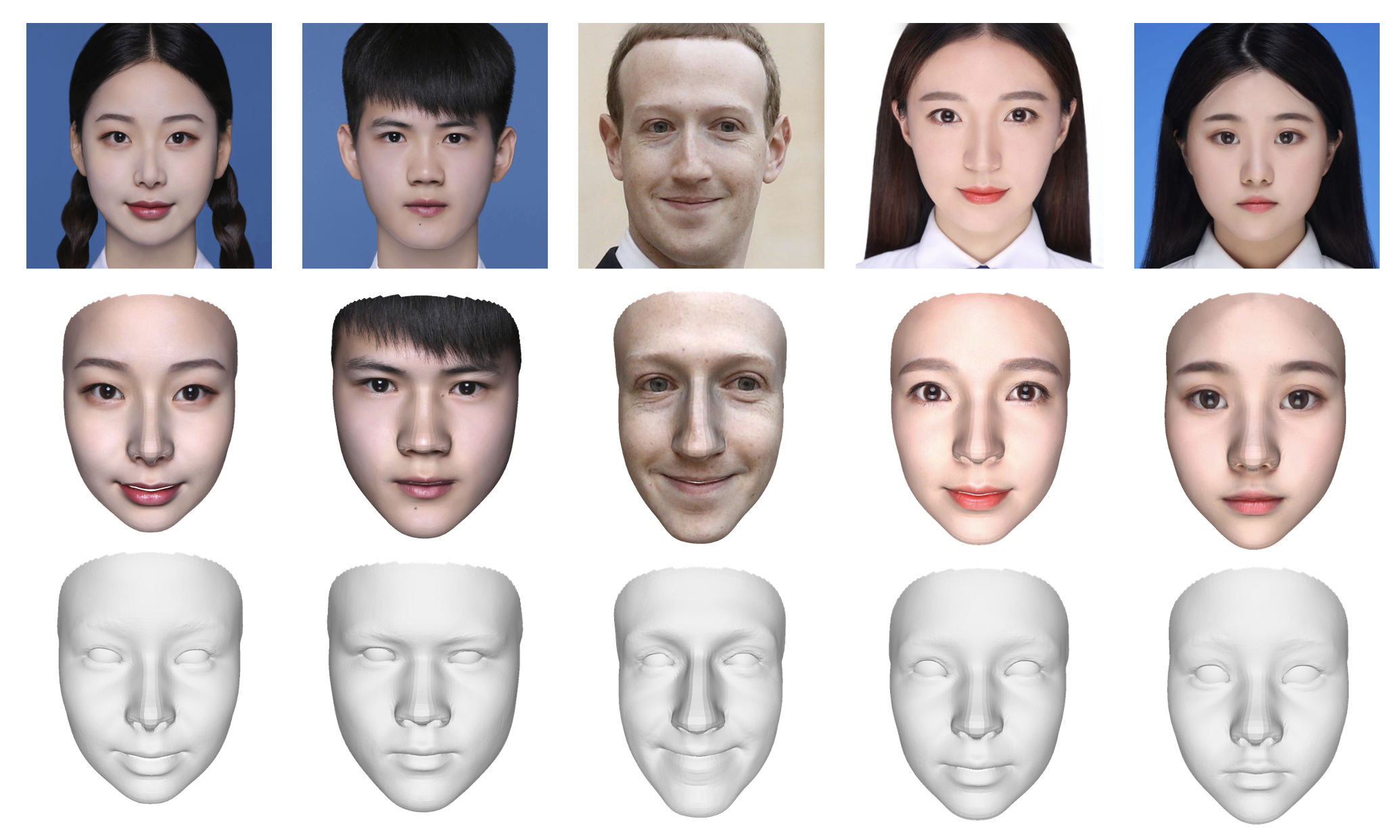}
\end{center}
   \caption{Realistic 3D face reconstruction results \YL{from ordinary portraits}. The upper row is the input image. The middle row and the bottom row are reconstructed models with and without texture respectively.}
\label{fig:real-world}
\end{figure}

\paragraph{Results on \YL{ordinary portrait images}.} Although we focus on cartoon face reconstruction, our method can also reconstruct high-quality realistic faces. Fig.~\ref{fig:real-world} shows examples of  single-view 3D face reconstruction \YL{from ordinary portraits} with our method.

\subsection{More results}

\begin{figure}[t]
\begin{center}
\includegraphics[width=1.0\linewidth]{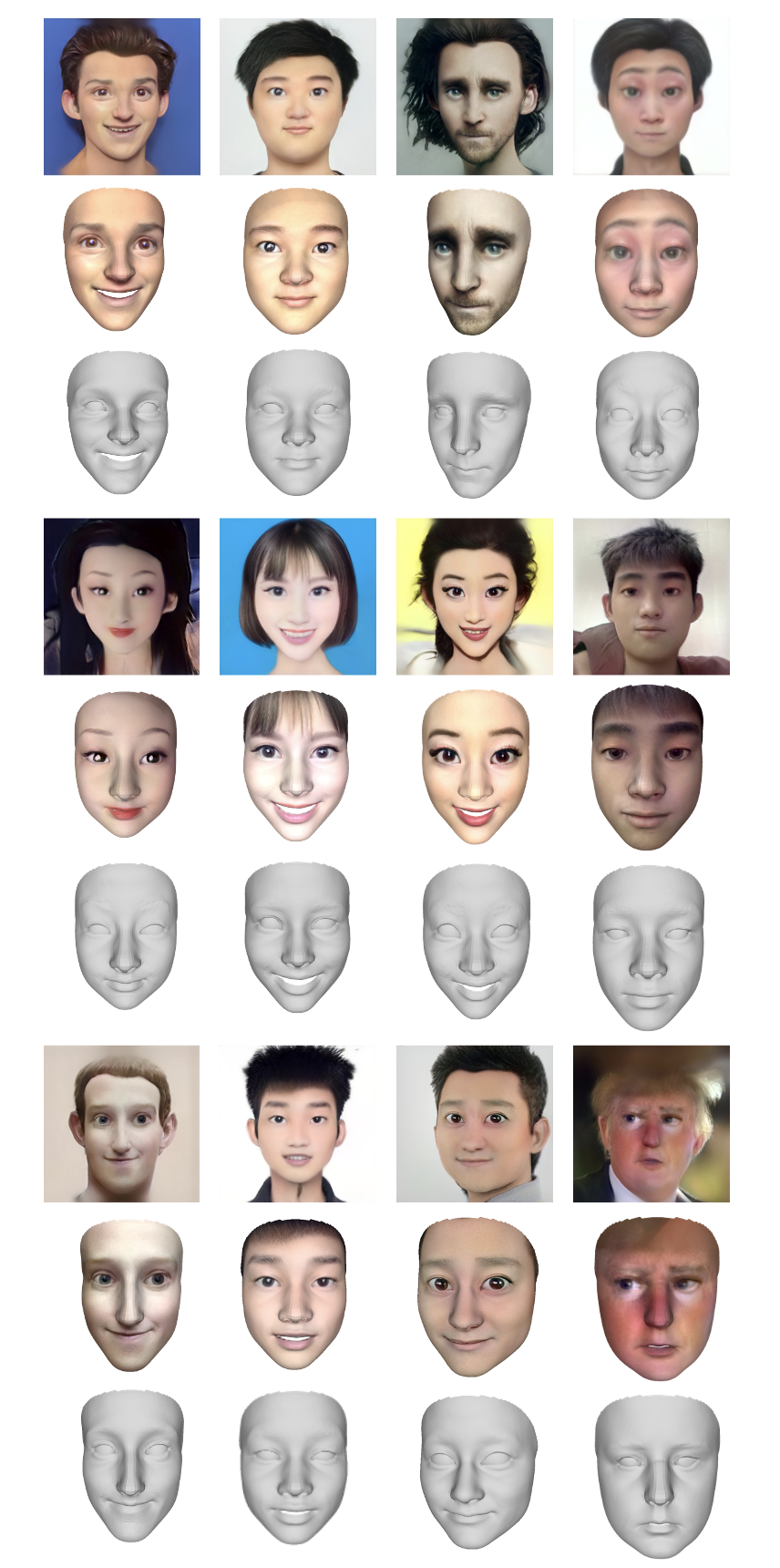}
\end{center}
   \caption{More visualization results. We conduct reconstruction on images with different styles. For \YL{every} three rows, the first row shows the input cartoon images, the second and third row show 3D models with and without texture.}
\label{fig:results}
\end{figure}

Fig.~\ref{fig:results} shows more visualization results on cartoon images with different styles. Our method is robust to exaggerated face parts like large eyes and unnatural face shapes.

\section{Conclusion}

In this paper, we introduce a novel pipeline to generate animatable 3D cartoon faces from a single real-world portrait. To achieve high-quality 3D cartoon faces, we propose a two-stage face reconstruction scheme. We generate semantic-preserving face \YL{rigs} with \YL{manually-created} models and expression transfer. Quantitative and Qualitative results show that our reconstruction \YL{achieves} high performance on accuracy, aesthetics and similarity. Furthermore, we show the capability of real-time animation of our model. Our pipeline can be applied to creating user 3D \YL{avatars} in VR/AR applications. Generating high-quality animatable 3D faces with various styles is a difficult task, and we would like to generalize our method to a larger range of styles in our future research agenda.

\bibliographystyle{spmpsci}
\bibliography{cvmbib} 

\begin{thebibliography}{10}
\providecommand{\url}[1]{{#1}}
\providecommand{\urlprefix}{URL }
\expandafter\ifx\csname urlstyle\endcsname\relax
  \providecommand{\doi}[1]{DOI~\discretionary{}{}{}#1}\else
  \providecommand{\doi}{DOI~\discretionary{}{}{}\begingroup
  \urlstyle{rm}\Url}\fi

\bibitem{alexander2010digital}
Alexander, O., Rogers, M., Lambeth, W., Chiang, J.Y., Ma, W.C., Wang, C.C.,
  Debevec, P.: The digital {Emily} project: Achieving a photorealistic digital
  actor.
\newblock IEEE Computer Graphics and Applications \textbf{30}(4), 20--31 (2010)

\bibitem{bas2016fitting}
Bas, A., Smith, W.A., Bolkart, T., Wuhrer, S.: Fitting a {3D} morphable model
  to edges: A comparison between hard and soft correspondences.
\newblock In: Asian Conference on Computer Vision, pp. 377--391. Springer
  (2016)

\bibitem{blanz2003reanimating}
Blanz, V., Basso, C., Poggio, T., Vetter, T.: Reanimating faces in images and
  video.
\newblock In: Computer graphics forum, vol.~22, pp. 641--650. Wiley Online
  Library (2003)

\bibitem{blanz2004statistical}
Blanz, V., Mehl, A., Vetter, T., Seidel, H.P.: A statistical method for robust
  {3D} surface reconstruction from sparse data.
\newblock In: Proceedings. 2nd International Symposium on 3D Data Processing,
  Visualization and Transmission, 2004. 3DPVT 2004., pp. 293--300. IEEE (2004)

\bibitem{blanz1999morphable}
Blanz, V., Vetter, T.: A morphable model for the synthesis of {3D} faces.
\newblock In: Proceedings of the 26th annual conference on Computer graphics
  and interactive techniques, pp. 187--194 (1999)

\bibitem{cai2021landmark}
Cai, H., Guo, Y., Peng, Z., Zhang, J.: Landmark detection and {3D} face
  reconstruction for caricature using a nonlinear parametric model.
\newblock Graphical Models \textbf{115}, 101,103 (2021)

\bibitem{cao2015real}
Cao, C., Bradley, D., Zhou, K., Beeler, T.: Real-time high-fidelity facial
  performance capture.
\newblock ACM Transactions on Graphics (ToG) \textbf{34}(4), 1--9 (2015)

\bibitem{casas2016rapid}
Casas, D., Feng, A., Alexander, O., Fyffe, G., Debevec, P., Ichikari, R., Li,
  H., Olszewski, K., Suma, E., Shapiro, A.: Rapid photorealistic blendshape
  modeling from {RGB-D} sensors.
\newblock In: Proceedings of the 29th International Conference on Computer
  Animation and Social Agents, pp. 121--129 (2016)

\bibitem{blender}
Community, B.O.: Blender - a {3D} modelling and rendering package.
\newblock Blender Foundation, Stichting Blender Foundation, Amsterdam (2018).
\newblock \urlprefix\url{http://www.blender.org}

\bibitem{deng2019accurate}
Deng, Y., Yang, J., Xu, S., Chen, D., Jia, Y., Tong, X.: Accurate {3D} face
  reconstruction with weakly-supervised learning: From single image to image
  set.
\newblock In: Proceedings of the IEEE/CVF Conference on Computer Vision and
  Pattern Recognition Workshops, pp. 0--0 (2019)

\bibitem{ekman1978facial}
Ekman, P., Friesen, W.V.: Facial action coding system.
\newblock Environmental Psychology \& Nonverbal Behavior  (1978)

\bibitem{feng2018joint}
Feng, Y., Wu, F., Shao, X., Wang, Y., Zhou, X.: Joint {3D} face reconstruction
  and dense alignment with position map regression network.
\newblock In: Proceedings of the European conference on computer vision (ECCV),
  pp. 534--551 (2018)

\bibitem{garrido2016reconstruction}
Garrido, P., Zollh{\"o}fer, M., Casas, D., Valgaerts, L., Varanasi, K.,
  P{\'e}rez, P., Theobalt, C.: Reconstruction of personalized {3D} face rigs
  from monocular video.
\newblock ACM Transactions on Graphics (TOG) \textbf{35}(3), 1--15 (2016)

\bibitem{guo2018cnn}
Guo, Y., Cai, J., Jiang, B., Zheng, J., et~al.: {CNN}-based real-time dense
  face reconstruction with inverse-rendered photo-realistic face images.
\newblock IEEE transactions on pattern analysis and machine intelligence
  \textbf{41}(6), 1294--1307 (2018)

\bibitem{hassner2013viewing}
Hassner, T.: Viewing real-world faces in {3D}.
\newblock In: Proceedings of the IEEE International Conference on Computer
  Vision, pp. 3607--3614 (2013)

\bibitem{hassner2006example}
Hassner, T., Basri, R.: Example based {3D} reconstruction from single {2D}
  images.
\newblock In: 2006 Conference on Computer Vision and Pattern Recognition
  Workshop (CVPRW'06), pp. 15--15. IEEE (2006)

\bibitem{hassner2015effective}
Hassner, T., Harel, S., Paz, E., Enbar, R.: Effective face frontalization in
  unconstrained images.
\newblock In: Proceedings of the IEEE conference on computer vision and pattern
  recognition, pp. 4295--4304 (2015)

\bibitem{ichim2015dynamic}
Ichim, A.E., Bouaziz, S., Pauly, M.: Dynamic {3D} avatar creation from
  hand-held video input.
\newblock ACM Transactions on Graphics (ToG) \textbf{34}(4), 1--14 (2015)

\bibitem{jourabloo2016large}
Jourabloo, A., Liu, X.: Large-pose face alignment via {CNN}-based dense {3D}
  model fitting.
\newblock In: Proceedings of the IEEE conference on computer vision and pattern
  recognition, pp. 4188--4196 (2016)

\bibitem{karras2019style}
Karras, T., Laine, S., Aila, T.: A style-based generator architecture for
  generative adversarial networks.
\newblock In: Proceedings of the IEEE/CVF conference on computer vision and
  pattern recognition, pp. 4401--4410 (2019)

\bibitem{kemelmacher2011face}
Kemelmacher-Shlizerman, I., Seitz, S.M.: Face reconstruction in the wild.
\newblock In: 2011 international conference on computer vision, pp. 1746--1753.
  IEEE (2011)

\bibitem{kim2017inversefacenet}
Kim, H., Zollh{\"o}fer, M., Tewari, A., Thies, J., Richardt, C., Theobalt, C.:
  Inversefacenet: Deep single-shot inverse face rendering from a single image.
\newblock arXiv preprint arXiv:1703.10956  (2017)

\bibitem{king2009dlib}
King, D.E.: Dlib-ml: A machine learning toolkit.
\newblock The Journal of Machine Learning Research \textbf{10}, 1755--1758
  (2009)

\bibitem{Laine2020diffrast}
Laine, S., Hellsten, J., Karras, T., Seol, Y., Lehtinen, J., Aila, T.: Modular
  primitives for high-performance differentiable rendering.
\newblock ACM Transactions on Graphics \textbf{39}(6) (2020)

\bibitem{lewis2014practice}
Lewis, J.P., Anjyo, K., Rhee, T., Zhang, M., Pighin, F.H., Deng, Z.: Practice
  and theory of blendshape facial models.
\newblock Eurographics (State of the Art Reports) \textbf{1}(8), 2 (2014)

\bibitem{li2010example}
Li, H., Weise, T., Pauly, M.: Example-based facial rigging.
\newblock Acm transactions on graphics (tog) \textbf{29}(4), 1--6 (2010)

\bibitem{liu2009semi}
Liu, J., Chen, Y., Miao, C., Xie, J., Ling, C.X., Gao, X., Gao, W.:
  Semi-supervised learning in reconstructed manifold space for {3D} caricature
  generation.
\newblock In: Computer Graphics Forum, vol.~28, pp. 2104--2116. Wiley Online
  Library (2009)

\bibitem{lugaresi2019mediapipe}
Lugaresi, C., Tang, J., Nash, H., McClanahan, C., Uboweja, E., Hays, M., Zhang,
  F., Chang, C.L., Yong, M.G., Lee, J., et~al.: Mediapipe: A framework for
  building perception pipelines.
\newblock arXiv preprint arXiv:1906.08172  (2019)

\bibitem{muslim2019knowledge}
Muslim, H.S.M., Khan, S.A., Hussain, S., Jamal, A., Qasim, H.S.A.: A
  knowledge-based image enhancement and denoising approach.
\newblock Computational and Mathematical Organization Theory \textbf{25},
  108--121 (2019)

\bibitem{paszke2019pytorch}
Paszke, A., Gross, S., Massa, F., Lerer, A., Bradbury, J., Chanan, G., Killeen,
  T., Lin, Z., Gimelshein, N., Antiga, L., et~al.: {PyTorch}: An imperative
  style, high-performance deep learning library.
\newblock Advances in neural information processing systems \textbf{32} (2019)

\bibitem{pawaskar2013expression}
Pawaskar, C., Ma, W.C., Carnegie, K., Lewis, J.P., Rhee, T.: Expression
  transfer: A system to build {3D} blend shapes for facial animation.
\newblock In: 2013 28th International Conference on Image and Vision Computing
  New Zealand (IVCNZ 2013), pp. 154--159. IEEE (2013)

\bibitem{pinkney2020resolution}
Pinkney, J.N., Adler, D.: Resolution dependent {GAN} interpolation for
  controllable image synthesis between domains.
\newblock arXiv preprint arXiv:2010.05334  (2020)

\bibitem{qiu20213dcaricshop}
Qiu, Y., Xu, X., Qiu, L., Pan, Y., Wu, Y., Chen, W., Han, X.: {3DCaricShop}: A
  dataset and a baseline method for single-view {3D} caricature face
  reconstruction.
\newblock In: Proceedings of the IEEE/CVF Conference on Computer Vision and
  Pattern Recognition, pp. 10,236--10,245 (2021)

\bibitem{ramamoorthi2001efficient}
Ramamoorthi, R., Hanrahan, P.: An efficient representation for irradiance
  environment maps.
\newblock In: Proceedings of the 28th annual conference on Computer graphics
  and interactive techniques, pp. 497--500 (2001)

\bibitem{romdhani2005estimating}
Romdhani, S., Vetter, T.: Estimating {3D} shape and texture using pixel
  intensity, edges, specular highlights, texture constraints and a prior.
\newblock In: 2005 IEEE Computer Society Conference on Computer Vision and
  Pattern Recognition (CVPR'05), vol.~2, pp. 986--993. IEEE (2005)

\bibitem{saito2019pifu}
Saito, S., Huang, Z., Natsume, R., Morishima, S., Kanazawa, A., Li, H.: {PIFu}:
  Pixel-aligned implicit function for high-resolution clothed human
  digitization.
\newblock In: Proceedings of the IEEE/CVF International Conference on Computer
  Vision, pp. 2304--2314 (2019)

\bibitem{sumner2004deformation}
Sumner, R.W., Popovi{\'c}, J.: Deformation transfer for triangle meshes.
\newblock ACM Transactions on graphics (TOG) \textbf{23}(3), 399--405 (2004)

\bibitem{vlasic2006face}
Vlasic, D., Brand, M., Pfister, H., Popovic, J.: Face transfer with multilinear
  models.
\newblock In: ACM SIGGRAPH 2006 Courses, pp. 24--es (2006)

\bibitem{wu2018alive}
Wu, Q., Zhang, J., Lai, Y.K., Zheng, J., Cai, J.: Alive caricature from {2D} to
  {3D}.
\newblock In: Proceedings of the IEEE Conference on Computer Vision and Pattern
  Recognition, pp. 7336--7345 (2018)

\bibitem{yu2021bisenet}
Yu, C., Gao, C., Wang, J., Yu, G., Shen, C., Sang, N.: {BiSeNet} v2: Bilateral
  network with guided aggregation for real-time semantic segmentation.
\newblock International Journal of Computer Vision \textbf{129}(11), 3051--3068
  (2021)

\bibitem{zhang2016joint}
Zhang, K., Zhang, Z., Li, Z., Qiao, Y.: Joint face detection and alignment
  using multitask cascaded convolutional networks.
\newblock IEEE signal processing letters \textbf{23}(10), 1499--1503 (2016)

\bibitem{zhang2021self}
Zhang, Y., Di, X., Zhang, B., Li, Q., Yan, S., Wang, C.: Self-supervised low
  light image enhancement and denoising.
\newblock arXiv preprint arXiv:2103.00832  (2021)

\bibitem{zhou20183d}
Zhou, J., Wu, H.T., Liu, Z., Tong, X., Guo, B.: {3D} cartoon face rigging from
  sparse examples.
\newblock The Visual Computer \textbf{34}(9), 1177--1187 (2018)

\bibitem{zhou2005large}
Zhou, K., Huang, J., Snyder, J., Liu, X., Bao, H., Guo, B., Shum, H.Y.: Large
  mesh deformation using the volumetric graph {Laplacian}.
\newblock In: ACM SIGGRAPH 2005 Papers, pp. 496--503 (2005)

\bibitem{zhu2016face}
Zhu, X., Lei, Z., Liu, X., Shi, H., Li, S.Z.: Face alignment across large
  poses: A {3D} solution.
\newblock In: Proceedings of the IEEE conference on computer vision and pattern
  recognition, pp. 146--155 (2016)

\bibitem{zhu2015high}
Zhu, X., Lei, Z., Yan, J., Yi, D., Li, S.Z.: High-fidelity pose and expression
  normalization for face recognition in the wild.
\newblock In: Proceedings of the IEEE conference on computer vision and pattern
  recognition, pp. 787--796 (2015)

\end{thebibliography}

\end{document}